\documentclass[astyle]{JWPublic}

\usepackage{amsmath}
\usepackage{amssymb}

\newcommand{\indep}{\perp\hspace{-0.21cm}\perp}
\newcommand{\given}{\operatorname{|}}
\newcommand{\Prob}{\operatorname{P}}

\begin{document}

\ContribChap{Introduction to Graphical Modelling}%
    {Marco Scutari ${}^1$ and Korbinian Strimmer ${}^2$ }%
    {${}^1$ Genetics Institute, University College London (UCL), London, UK \\
    ${}^2$ Institute for Medical Informatics, Statistics and Epidemiology (IMISE),
    University of Leipzig, Germany}

The aim of this chapter is twofold. In the first part (Sections \ref{sec:definitions},
\ref{sec:learning} and \ref{sec:inference}) we will provide a brief overview of
the mathematical and statistical foundations of graphical models, along with their
fundamental properties, estimation and basic inference procedures. In particular
we will develop Markov networks (also known as Markov random fields) and Bayesian
networks, which are the subjects of most past and current literature on graphical
models. In the second part (Section \ref{sec:applications}) we will review some
applications of graphical models in systems biology.

\section{Graphical Structures and Random Variables}
\label{sec:definitions}

Graphical models are a class of statistical models which combine the rigour of
a probabilistic approach with the intuitive representation of relationships
given by graphs. They are composed by two parts:
\begin{enumerate}
  \item a set $\mathbf{X} = \{X_1, X_2, \ldots, X_p\}$ of \textit{random variables}
    describing the quantities of interest. The statistical distribution of $\mathbf{X}$
    is called the \textit{global distribution} of the data, while the components it
    factorises into are called \textit{local distributions}.
  \item a \textit{graph} $\mathcal{G} = (\mathbf{V}, E)$ in which each \textit{vertex}
    $v \in \mathbf{V}$, also called a \textit{node}, is associated with one of the
    random variables in $\mathbf{X}$ (they are usually referred to interchangeably).
    \textit{Edges} $e \in E$, also called \textit{links}, are used to express the
    \textit{dependence structure} of the data (the set of dependence relationships
    among the variables in $\mathbf{X}$) with different semantics for
    \textit{undirected graphs} \citep{diestel} and \textit{directed acyclic graphs}
    \citep{digraphs}.
\end{enumerate}

The scope of this class of models and the versatility of its definition are well
expressed by \citet{pearl} in his seminal work `Probabilistic Reasoning in
Intelligent Systems: Networks of Plausible Inference':

\begin{quote}
  Graph representations meet our earlier requirements of explicitness, saliency, and
  stability. The links in the graph permit us to express directly and quantitatively
  the dependence relationships, and the graph topology displays these relationships
  explicitly and preserves them, under any assignment of numerical parameters.
\end{quote}

The nature of the link outlined above between the dependence structure of the data
and its graphical representation is given again by \citet{pearl} in terms of
\textit{conditional independence} (denoted with $\indep_P$) and \textit{graphical
separation} (denoted with $\indep_G$).

\begin{definition}
\label{defn:maps}
  A graph $\mathcal{G}$ is a dependency map (or \mbox{D-map}) of the probabilistic
  dependence structure $P$ of $\mathbf{X}$ if there is a one-to-one correspondence
  between the random variables in $\mathbf{X}$ and the nodes $\mathbf{V}$ of 
  $\mathcal{G}$, such that for all disjoint subsets $\mathbf{A}$, $\mathbf{B}$,
  $\mathbf{C}$ of $\mathbf{X}$ we have
  \begin{equation}
    \mathbf{A} \indep_P \mathbf{B} \given \mathbf{C} \Longrightarrow
    \mathbf{A} \indep_G \mathbf{B} \given \mathbf{C}.
  \end{equation}
  Similarly, $\mathcal{G}$ is an independency map (or \mbox{I-map}) of $P$ if
  \begin{equation}
    \mathbf{A} \indep_P \mathbf{B} \given \mathbf{C} \Longleftarrow
    \mathbf{A} \indep_G \mathbf{B} \given \mathbf{C}.
  \end{equation}
  $\mathcal{G}$ is said to be a perfect map of $P$ if it is both
  a \mbox{D-map} and an \mbox{I-map}, that is
  \begin{equation}
    \mathbf{A} \indep_P \mathbf{B} \given \mathbf{C} \Longleftrightarrow
    \mathbf{A} \indep_G \mathbf{B} \given \mathbf{C},
  \end{equation}
  and in this case $P$ is said to be isomorphic to $\mathcal{G}$.
\end{definition}

Note that this definition does not depend on a particular characterisation of
graphical separation, and therefore on the type of graph used in the graphical
model. In fact both \textit{Markov networks} \citep{whittaker} and \textit{Bayesian
networks}  \citep{pearl}, which are by far the two most common classes of 
graphical models treated in literature, are defined as \textit{minimal I-maps}
even though the former use undirected graphs an the latter use directed acyclic
graphs. Minimality requires that, if the dependence structure $P$ of $\mathbf{X}$
can be expressed by multiple graphs, we must use the one with the minimum number
of edges; if any further edge is removed then the graph is no longer an I-map of
$P$. Being an I-map guarantees that two disjoint sets of nodes $\mathbf{A}$ and
$\mathbf{B}$ found to be separated by $\mathbf{C}$ in the graph (according to
the characterisation of separation for that type of graph) correspond to 
independent sets of variables. However, this does not mean that every conditional
independence relationship present in $P$ is reflected in the graph; this is true
only if the graph is also assumed to be a dependency map, making it a perfect map
of $P$. 

In Markov networks graphical separation (which is called \textit{undirected
separation} or \textit{u-separation} in \citet{castillo}) is easily defined
due to the lack of direction of the links.

\begin{definition}
\label{defn:usep}
  If $\mathbf{A}$, $\mathbf{B}$ and $\mathbf{C}$ are three disjoint subsets
  of nodes in an undirected graph $\mathcal{G}$, then $\mathbf{C}$ is
  said to separate $\mathbf{A}$ from $\mathbf{B}$, denoted
  $\mathbf{A} \indep_G \mathbf{B} \given \mathbf{C}$, if every path between
  a node in $\mathbf{A}$ and a node in $\mathbf{B}$ contains at least one
  node in $\mathbf{C}$.
\end{definition}

In Bayesian networks separation takes the name of \textit{directed separation}
(or \mbox{\textit{d-separation}}) and is defined as follows \citep{korb}.

\begin{definition}
\label{defn:dsep}
  If $\mathbf{A}$, $\mathbf{B}$ and $\mathbf{C}$ are three disjoint subsets
  of nodes in a directed acyclic graph $\mathcal{G}$, then $\mathbf{C}$ is
  said to \mbox{d-separate} $\mathbf{A}$ from $\mathbf{B}$, denoted
  $\mathbf{A} \indep_G \mathbf{B} \given \mathbf{C}$, if along every path
  between a node in $\mathbf{A}$ and a node in $\mathbf{B}$ there is a
  node $v$ satisfying one of the following two conditions:
  \begin{enumerate}
    \item $v$ has converging edges (i.e. there are two edges pointing to $v$
      from the adjacent nodes in the path) and none of $v$ or its descendants
      (i.e. the nodes that can be reached from $v$) are in $\mathbf{C}$.
    \item $v$ is in $\mathbf{C}$ and does not have converging edges.
  \end{enumerate}
\end{definition}

\begin{figure}[t]
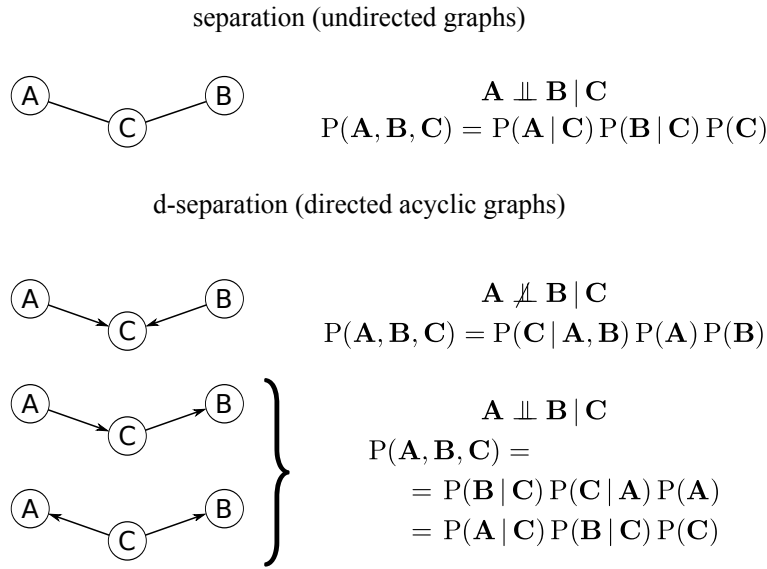

  \eps{images/fig1}{10}
  \caption{Graphical separation, conditional independence and probability
    factorisation for some simple undirected and directed acyclic graphs.
    The undirected graph is a simple 3 node chain, while the directed
    acyclic graphs are the \textit{converging}, \textit{serial} and
    \textit{diverging connections} (collectively known as
    \textit{fundamental connections} in the theory of Bayesian networks).}
  \label{fig:dsep}
\end{figure}

A simple application of these definitions is illustrated in Figure \ref{fig:dsep}.
We can see that in the undirected graph on the top $A$ and $B$ are separated by
$C$, because there is no edge between $A$ and $B$ and the path that connects them
contains $C$; so we can conclude that $A$ is independent from $B$ given $C$ according
to Definition \ref{defn:usep}. As for three directed acyclic graphs, which are 
called the \textit{converging}, \textit{serial} and \textit{diverging connections},
we can see that only the last two satisfy the conditions stated in Definition 
\ref{defn:dsep}. In the converging connection $C$ has two incoming edges (which
violates the second condition) and is included in the set of nodes we are
conditioning on (which violates the first condition). Therefore we can conclude
that $C$ does not \mbox{d-separate} $A$ and $B$ and that according to the 
definition of \mbox{I-map} we can not say that $A$ is independent from $B$
given $C$.

A fundamental result descending from the definitions of separation and 
\mbox{d-separation} is the \textit{Markov property} (or \textit{Markov condition}),
which defines the decomposition of the global distribution of the data into
a set of local distributions. For Bayesian networks it is related to the chain
rule of probability \citep{korb}; it takes the form
\begin{align}
\label{eqn:parents}
  \Prob(\mathbf{X}) &= \prod_{i=1}^p \Prob(X_i \given \Pi_{X_i})&
  & \text{for discrete data and}\\
  f(\mathbf{X}) &= \prod_{i=1}^p f(X_i \given \Pi_{X_i})&
  & \text{for continuous data,}
\end{align}
so that each local distribution is associated with a single node $X_i$ and
depends only on the joint distribution of its parents $\Pi_{X_i}$. This
decomposition holds for any Bayesian network, regardless of its graph structure.
In Markov networks on the other hand local distributions are associated with
the \textit{cliques} (maximal subsets of nodes in which each element is adjacent
to all the others) $\mathbf{C}_1$, $\mathbf{C}_2$, $\ldots$, $\mathbf{C}_k$
present in the graph; so
\begin{align}
\label{eqn:cliques}
  \Prob(\mathbf{X}) &= \prod_{i=1}^k \psi_i(\mathbf{C}_i)&
  & \text{for discrete data and}\\
  f(\mathbf{X}) &= \prod_{i=1}^k \psi_i(\mathbf{C}_i)&
  & \text{for continuous data.}
\end{align}
The functions $\psi_1, \psi_2, \ldots, \psi_k$ are called \textit{Gibbs'
potentials} \citep{pearl}, \textit{factor potentials} \citep{castillo} or simply
\textit{potentials}, and are non-negative functions representing the relative
mass of probability of each clique. They are proper probability or density 
functions only when the graph is \textit{decomposable} or \textit{triangulated},
that is when it contains no induced cycles other than triangles. With any other
type of graph inference becomes very hard, if possible at all, because $\psi_1, \psi_2,
\ldots, \psi_k$ have no direct statistical interpretation. Decomposable graphs
are also called \textit{chordal} \citep{diestel} because any cycle of length at
least four has a chord (a link between two nodes in a cycle that is not contained
in the cycle itself). In this case the global distribution factorises again 
according to the chain rule and can be written as
\begin{align}
  \Prob(\mathbf{X}) &= \frac{\prod_{i=1}^k \Prob(\mathbf{C}_i)}{\prod_{i=1}^k \Prob(\mathbf{S}_i)}&
  & \text{for discrete data and}\\
  f(\mathbf{X}) &= \frac{\prod_{i=1}^k f(\mathbf{C}_i)}{\prod_{i=1}^k f(\mathbf{S}_i)}&
  & \text{for continuous data,}
\end{align}
where $\mathbf{S}_i$ are the nodes of $\mathbf{C}_i$ which are also part
of any other clique up to $\mathbf{C}_{i-1}$ \citep{pearl}.

A trivial application of these factorisations is illustrated again in Figure
\ref{fig:dsep}. The Markov network is composed by two cliques, $C_1 = \left\{A, C\right\}$
and $C_2 = \left\{B, C\right\}$, separated by $R_1 = \left\{C\right\}$. Therefore
according to Equation \ref{eqn:cliques} we have
\begin{equation}
\Prob\left(\mathbf{X}\right)  
  = \frac{\Prob\left(A, C\right)\Prob\left(B, C\right)}{\Prob\left(C\right)}
  = \Prob\left(A \given C\right)\Prob\left(B \given C\right) \Prob\left(C\right).
\end{equation}
In the Bayesian networks we can see that the decomposition of the global distribution
results in three local distributions, one for each node. Each local distribution is
conditional on the set of parents of that particular node. For example, in the converging
connection we have that $\Pi_A = \left\{\varnothing\right\}$,
$\Pi_B = \left\{\varnothing\right\}$ and $\Pi_C = \left\{A, B\right\}$, so according
to Equation \ref{eqn:parents} the correct factorisation is
\begin{equation}
  \Prob\left(\mathbf{X}\right) = \Prob\left(A\right) \Prob\left(B\right) \Prob\left(C \given A, B\right).
\end{equation}
On the other hand, in the serial connection we have that $\Pi_A = \left\{\varnothing\right\}$,
$\Pi_B = \left\{C\right\}$ and $\Pi_C = \left\{A\right\}$, so
\begin{equation}
  \Prob\left(\mathbf{X}\right) = \Prob\left(A\right) \Prob\left(C \given A\right) \Prob\left(B \given C\right).
\end{equation}
The diverging connection can be shown to result in the same factorisation, even
though the nodes have different sets of parents than in the serial connection.

\begin{figure}
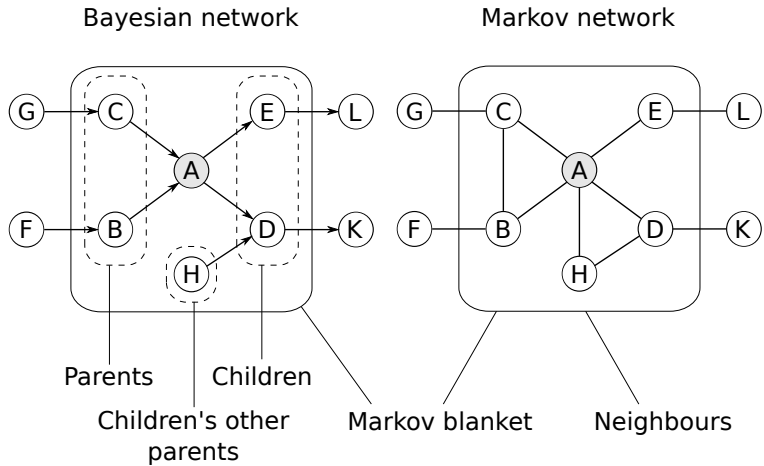

  \eps{images/fig2}{10}
  \caption{The Markov blanket of the node $A$ in a Bayesian network (on the left)
    and in the corresponding Markov network given by its \textit{moral graph}
    (on the right). The two graphs express the same dependence structure, so the
    Markov blanket of $A$ is the same.}
  \label{fig:blanket}
\end{figure}

Another fundamental result descending from the link between graphical separation
and probabilistic independence is the definition of the \textit{Markov blanket}
\citep{pearl} of a node $X_i$, the set that completely separates $X_i$ from the
rest of the graph. Generally speaking it is the set of nodes that includes all
the knowledge needed to do inference on $X_i$, from estimation to hypothesis
testing to prediction, because all the other nodes are conditionally independent
from $X_i$ given its Markov blanket. In Markov networks the Markov blanket coincides
with the neighbours of $X_i$ (all the nodes that are connected to $X_i$ by an
edge); in Bayesian networks it is the union of the children of $X_i$, its parents,
and its children's other parents (see Figure \ref{fig:blanket}). In both classes
of models the usefulness of Markov blankets is limited by the sparseness of the
network. If edges are few compared to the number of nodes the interpretation
of each Markov blanket becomes a useful tool in understanding and predicting
the behaviour of the data.

The two characterisations of graphical separation and of the Markov properties
presented above do not appear to be closely related, to the point that these two
classes of graphical models seem to be very different in construction and
interpretation. There are indeed dependency models that have an undirected perfect
map but not a directed acyclic one, and vice versa (see \citet{pearl}, pages 
126 -- 127 for a simple example of a dependency structure that cannot be represented
as a Bayesian network). However, it can be shown \citep{pearl,castillo} that every
dependency structure that can be expressed by a decomposable graph can be modelled
both by a Markov network and a Bayesian network. This is clearly the case for the
small networks shown in Figure \ref{fig:blanket}, as the undirected graph obtained
from the Bayesian network by \textit{moralisation} (connecting parents which share
a common child) is decomposable. It can also be shown that every dependency model
expressible by an undirected graph is also expressible by a directed acyclic graph,
with the addition of some auxiliary nodes. These two results indicate that there is
a significant overlap between Markov and Bayesian networks, and that in many
cases both can be used to the same effect.

\section{Learning Graphical Models}
\label{sec:learning}

Fitting graphical models is called \textit{learning}, a term borrowed from expert
systems and artificial intelligence theory, and in general requires a two-step
process.

The first step consists in finding the graph structure that encodes the
conditional independencies present in the data. Ideally it should coincide with
the minimal \mbox{I-map} of the global distribution, or it should at least identify a
distribution as close as possible to the correct one in the probability space.
This step is called \textit{network structure} or simply \textit{structure
learning} \citep{korb, koller}, and is similar in approaches and terminology to
model selection procedures for classical statistical models.

The second step is called \textit{parameter learning} and, as the name suggests,
deals with the estimation of the parameters of the global distribution. This
task can be easily reduced to the estimation of the parameters of the local 
distributions because the network structure is known from the previous step.

Both structure and parameter learning are often performed using a combination
of numerical algorithms and prior knowledge on the data. Even though significant
progress have been made on performance and scalability of learning algorithms,
an effective use of prior knowledge and relevant theoretical results can still
speed up the learning process severalfold and improve the accuracy of the
resulting model. Such a boost has been used in the past to overcome the
limitations on computational power, leading to the development of the so-called
\textit{expert systems} (for real-world examples see the MUNIN \citep{munin},
ALARM \citep{alarm} and Hailfinder \citep{hailfinder} networks); it can still
be used today to tackle larger and larger problems and obtain reliable results.

\subsection{Structure learning}
\label{sec:structlearn}

Structure learning algorithms have seen a steady development over the past two
decades thanks to the increased availability of computational power and the
application of many results from probability, information and optimisation
theory. Despite the (sometimes confusing) variety of theoretical backgrounds
and terminology they can all be traced to only three approaches: constraint-based,
score-based and hybrid.

\textit{Constraint-based algorithms}  use statistical tests to learn conditional
independence relationships (called \textit{constraints} in this setting) from
the data and assume that the graph underlying the probability distribution is a
perfect map to determine the correct network structure. They have been developed
originally for Bayesian networks, but have been recently applied to Markov
networks as well (see for example the \textit{Grow-Shrink} algorithm 
\citep{mphd,gsmn}, which works with minor modifications in both cases). Their
main limitations are the lack of control of either the \textit{family-wise error
rate} \citep{fwer} or the \textit{false discovery rate} \citep{fdr} and the
additional assumptions needed by the tests themselves, which are often
asymptotic and with problematic regularity conditions.

\textit{Score-based algorithms} are closer to model selection techniques
developed in classical statistics and information theory. Each candidate
network is assigned a score reflecting its goodness of fit, which is then
taken as an objective function to maximise. Since the number of both undirected
graphs and directed acyclic graphs grows more than exponentially in the number
of nodes \citep{harary} an exhaustive search is not feasible in all but the
most trivial cases. This has led to an extensive use of \textit{heuristic
optimisation algorithms}, from local search (starting from an initial network
and changing one edge at a time) to genetic algorithms \citep{norvig}.
Convergence to a global maximum however is not guaranteed, as they can get stuck
into a local maximum because of the noise present in the data or a poor choice
in the tuning parameters of the score function.

\textit{Hybrid algorithms} use both statistical tests and score functions,
combining the previous two families of algorithms. The general approach is
described for Bayesian networks in \citet{sparse}, and has proved one of the
top performers up to date in \citet{mmhc}. Conditional independence tests are
used to learn at least part of the conditional independence relationships from
the data, thus restricting the search space for a subsequent score-based search.
The latter determines which edges are actually present in the graph and, in the
case of Bayesian networks, their direction.

\begin{figure}
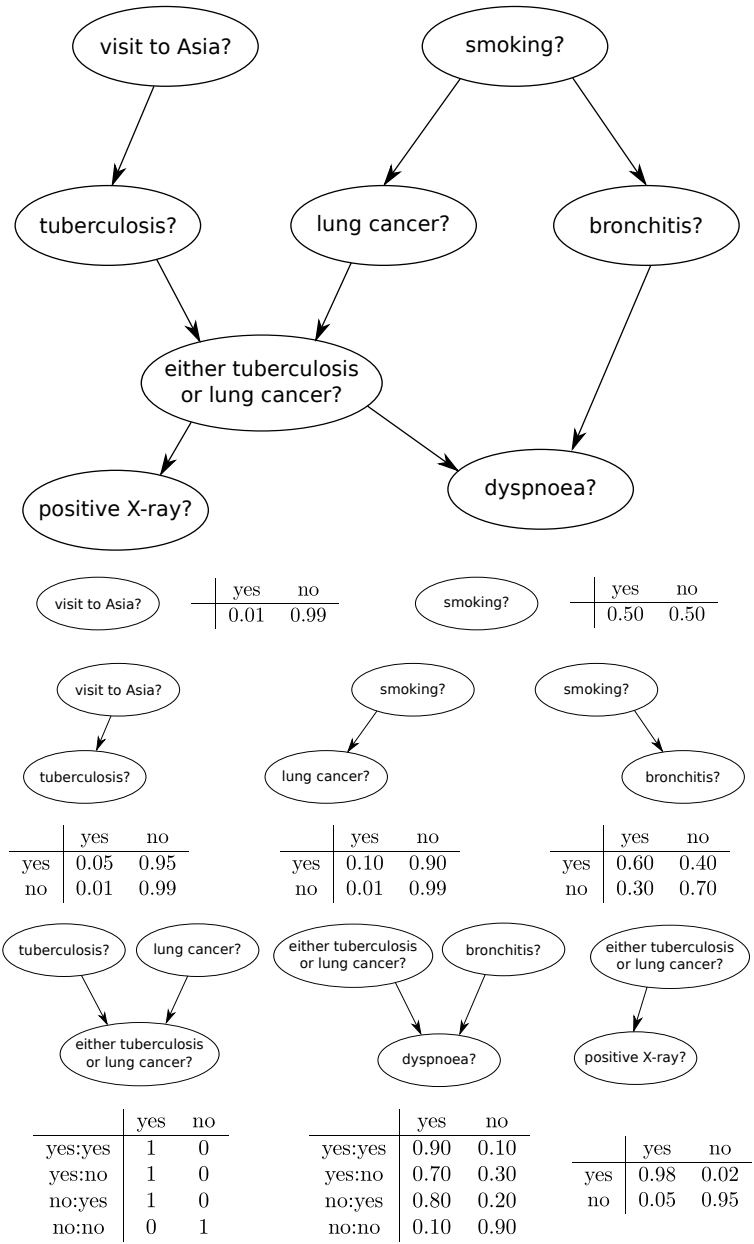

  \eps{images/fig3}{10}
  \caption{Factorisation of the ASIA Bayesian network from \citet{asia} into local 
    distributions, each with his own conditional probability table. Each row 
    contains the probabilities conditional on a particular configuration of 
    parents.}
  \label{fig:asia}
\end{figure}

All these structure learning algorithms operate under a set of common assumptions,
which are similar for Bayesian and Markov networks:
\begin{itemize}
  \item there must be a one-to-one correspondence between the nodes of the graph
    and the random variables included in the model; this means in particular that
    there must not be multiple nodes which are functions of a single variable.
  \item there must be no unobserved (also called \textit{latent} or \textit{hidden}) 
    variables that are parents of an observed node in a Bayesian network; otherwise
    only part of the dependency structure can be observed, and the model is likely
    to include spurious edges. Specific algorithms have been developed for this
    particular case, typically based on Bayesian posterior distributions or the
    EM algorithm \citep{dempster}; see for example \citet{hidden3}, \citet{hidden1}
    and \citet{hidden2}.
  \item all the relationships between the variables in the network must be
    conditional independencies, because they are by definition the only ones that
    can be expressed by graphical models.
  \item every combination of the possible values of the variables in $\mathbf{X}$
    must represent a valid, observable (even if really unlikely) event. This
    assumption implies a strictly positive global distribution, which is needed
    to have uniquely determined Markov blankets and, therefore, a uniquely 
    identifiable model. Constraint-based algorithms work even when this is not
    true, because the existence of a perfect map is also a sufficient condition
    for the uniqueness of the Markov blankets \citep{pearl}.
\end{itemize}
Some additional assumptions are needed to properly define the global distribution
of $\mathbf{X}$ and to have a tractable, closed-form decomposition:
\begin{figure}[b]
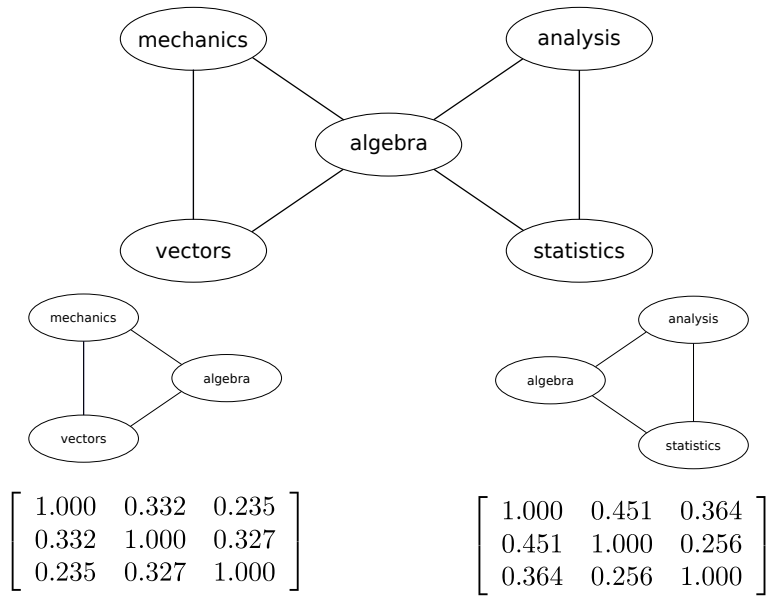

  \eps{images/fig3b}{10}
  \caption{The MARKS Graphical Gaussian network from \citet{edwards} and its 
    decomposition into cliques, the latter characterised by their partial 
    correlation matrices.}
  \label{marks}
\end{figure}
\begin{itemize}
  \item observations must be stochastically independent. If some form of temporal or
    spatial dependence is present it must be specifically accounted for in the
    definition of the network, as in \textit{dynamic Bayesian networks} \citep{koller}.
    They will be covered in Section \ref{sec:dbns} and Chapter 12.
  \item if all the random variables in $\mathbf{X}$ are discrete or categorical
    both the global and the local distributions are assumed to be \textit{multinomial}.
    This is by far the most common assumption in literature, at least for Bayesian
    networks, because of its strong ties with the analysis of contingency tables
    \citep{agresti, bishop} and because it allows an easy representation of local
    distributions as \textit{conditional probability tables} (see Figure \ref{fig:asia}).
  \item if on the other hand all the variables in $\mathbf{X}$ are continuous
    the global distribution is usually assumed to follow a \textit{multivariate
    Gaussian distribution}, and the local distributions are either \textit{univariate}
    or \textit{multivariate Gaussian distributions}. This assumption defines a 
    subclass of graphical models called \textit{graphical Gaussian models} (GGMs),
    which overlaps both Markov \citep{whittaker} and Bayesian networks \citep{neapolitan}.
    A classical example from \citet{edwards} is illustrated in Figure \ref{marks}.
  \item if both continuous and categorical variables are present in the data there
    are three possible choices: assuming a \textit{mixture} or \textit{conditional
    Gaussian distribution} \citep{bottcher,edwards}, discretising continuous
    attributes \citep{discretizing} or using a nonparametric approach \citep{mercer}.
\end{itemize}
The form of the probability or density function chosen for the local distributions
determines which score functions (for score-based algorithms) or conditional
independence tests (for constraint-based algorithms) can be used by structure 
learning algorithms. Common choices for conditional independence tests are:
\begin{itemize}
  \item \textit{discrete data}: Pearson's $\chi^2$ and the $G^2$ tests 
    \citep{agresti,edwards}, either as asymptotic or permutation tests. The $G^2$
    test is actually a log-likelihood ratio test \citep{lehmann} and is equivalent
    to mutual information tests \citep{itheory} up to a constant.
  \item \textit{continuous data}: Student's $t$, Fisher's $Z$ and the log-likelihood
     ratio tests based on partial correlation coefficients \citep{neapolitan,legendre}, 
     again either as asymptotic or permutation tests. The log-likelihood ratio
     test is equivalent to the corresponding mutual information test as before.
\end{itemize}
Score functions commonly used in both cases are penalised likelihood scores such
as the \textit{Akaike} and \textit{Bayesian Information criteria} (AIC and BIC, 
see \citet{akaike} and \citet{schwarz} respectively), posterior densities such
as the \textit{Bayesian Dirichlet} and \textit{Gaussian equivalent} scores 
(BDe and BGe, see \citet{heckerman} and \citet{heckerman2} respectively) and 
entropy-based measures such as the \textit{Minimum Description Length} (MDL)
by \citet{rissanen}.

The last important property of structure learning algorithms, one that sometimes
is not explicitly stated, is their inability to discriminate between \textit{score
equivalent} Bayesian networks \citep{chickering}. Such models have the same 
\textit{skeleton} (the undirected graph resulting from ignoring the direction
of every edge) and the same \mbox{\textit{v-structures}} (another name for the
converging connection illustrated in Figure \ref{fig:dsep}), and therefore they
encode the same conditional independence relationships because every 
\mbox{d-separation} statement that is true for one of them also holds for all
the others. 

\begin{figure}[b]
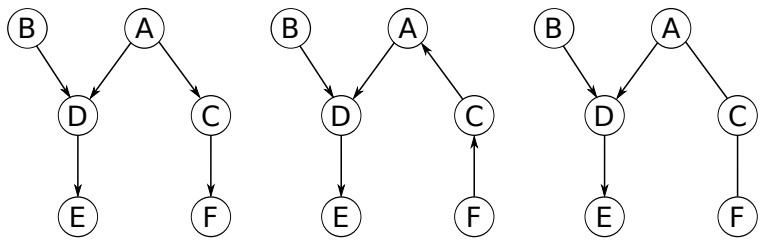

  \eps{images/fig4}{10}
  \caption{Two score equivalent Bayesian networks (on the left and in the
    middle) and the partially directed graph representing the equivalence
    class they belong to (on the right). Note that the direction of the edge 
    $\mathrm{D} \rightarrow \mathrm{E}$ is set because its reverse
    $\mathrm{E} \rightarrow \mathrm{D}$ would introduce two additional
    \mbox{v-structures} in the graph; for this reason it is called a 
    \textit{compelled edge} \citep{pearl}.}
  \label{fig:equiv}
\end{figure}

This characterisation implies a partitioning of the space of the possible networks
into a set of equivalence classes whose elements are all \mbox{I-maps} of the
same probability distribution. The elements of each of those equivalence
classes are indistinguishable from each other without additional information, 
such as a non-uniform prior distribution; their factorisations into local 
distributions are equivalent. Statistical tests and almost all score functions 
(which are in turn called \textit{score equivalent functions}), including 
those detailed above, are likewise unable to choose one model over an 
equivalent one. This means that learning algorithms, which base their decisions
on these very tests and scores, are only able to learn which equivalence class
the minimal \mbox{I-map} of the dependence structure belongs to. They are 
usually not able to uniquely determine the direction of all the edges present
in the network, which is then represented as a partially directed graph (see
Figure \ref{fig:equiv}).

\subsection{Parameter learning}

Once the structure of the network has been learned from the data the task of
estimating and updating the parameters of the global distribution is greatly
simplified by the application of the Markov property.

Local distributions in practise involve only a small number of variables; 
furthermore their dimension usually does not scale with the size
of $\mathbf{X}$ (and is often assumed to be bounded by a constant when computing
the computational complexity of algorithms), thus avoiding the so called
\textit{curse of dimensionality}. This means that each local distribution has a
comparatively small number of parameters to estimate from the sample, and that
estimates are more accurate due to the better ratio between the size of parameter
space and the sample size.

The number of parameters needed to uniquely identify the global distribution, which
is the sum of the number of parameters of the local ones, is also reduced because
the conditional independence relationships encoded in the network structure fix 
large parts of the parameter space. For example in graphical Gaussian models 
partial correlation coefficients involving (conditionally) independent variables
are equal to zero by definition, and joint frequencies factorise into marginal ones
in multinomial distributions.

However, parameter estimation is still problematic in many situations. For example
it is increasingly common to have sample size much smaller than the number of
variables included in the model; this is typical of microarray data, which have
a few tens or hundreds observations and thousands of genes. Such a situation, which
is called "small $n$, large $p$", leads to estimates with high variability unless
particular care is taken both in structure and parameter learning 
\citep{roverato,SS05c,elemstatlearn}.

Dense networks, which have a high number of edges compared their nodes, represent
another significant challenge. Exact inference quickly becomes unfeasible as the
number of nodes increases, and even approximate procedures based on Monte Carlo
simulations and bootstrap resampling require large computational resources
\citep{koller,korb}. Numerical problems stemming from floating point approximations
\citep{goldberg} and approximate numerical algorithms (such as the ones used in
matrix inversion and eigenvalue computation) should also be taken into account.

\section{Inference on Graphical Models}
\label{sec:inference}

Inference procedures for graphical models focus mainly on \textit{evidence
propagation} and model validation, even though other aspects such as 
robustness \citep{robustbn} and sensitivity analysis \citep{sensbn} have 
been studied for specific settings.

Evidence propagation (another term borrowed from expert systems literature)
studies the impact of new evidence and beliefs on the parameters of the model.
For this reason it is also referred to as \textit{belief propagation} or 
\textit{belief updating} \citep{pearl,castillo}, and has a clear Bayesian
interpretation in terms of posterior and conditional probabilities. The structure
of the network is usually considered fixed, thus allowing a scalable and efficient
updating of the model through its decomposition into local distributions.

In practise new evidence is introduced by either altering the relevant parameters
(\textit{soft evidence}) or setting one or more variables to a fixed value 
(\textit{hard evidence}). The former can be thought of as a model revision or 
parameter tuning process, while the latter is carried out by conditioning the 
behaviour of the network on the values of some nodes. The process of computing
such conditional probabilities is also known as \textit{conditional probability query}
on a set of \textit{query nodes} \citep{koller}, and can be performed with
a wide selection of exact and approximate inference algorithms. Two classical
examples of exact algorithms are variable elimination (optionally applied to
the \textit{clique tree} form of the network) and Kim and Pearl's Message Passing
algorithm. Approximate algorithms on the other hand rely on various forms of 
Monte Carlo sampling such as forward sampling (also called \textit{logic sampling}
for Bayesian networks), likelihood-weighted sampling and importance sampling.
Markov Chain Monte Carlo methods such as Gibbs sampling are also widely used
\citep{korb}.

Model validation on the other hand deals with the assessment of the performance
of a graphical model when dealing with new or existing data. Common measures 
are the goodness-of-fit scores cited in the Section \ref{sec:structlearn} or
any appropriate loss measure such as misclassification error (for discrete data)
and the residual sum of squares (for continuous data). Their estimation is 
usually carried out using either a separate testing data set or cross validation
\citep{cvdbn,koller} to avoid negatively biased results.

Another nontrivial problem is to determine the confidence level for particular
structural features. In \citet{friedman} this is accomplished by learning
a large number of Bayesian networks from bootstrap samples drawn from the 
original data set and estimating the empirical frequency of the features of
interest. \citet{mvber09} has recently extended this approach to obtain 
some univariate measures of variability and perform some basic hypothesis
testing. Both techniques can be applied to Markov networks with little
to no change. \citet{features} on the other hand used a non-uniform prior
distribution on the space of the possible structures to compute the exact
marginal posterior distribution of the features of interest.

\section{Application of Graphical Models in Systems Biology}
\label{sec:applications}

In \textit{systems biology} graphical models are employed to describe and to 
identify interdependencies among genes and gene products, with the eventual aim 
to better understand the molecular mechanisms of the cell. In \textit{medical 
systems biology} the specific focus lies on disease mechanisms mediated by 
changes in the network structure. For example, a general assumption in cancer 
genomics is that there are different active pathways in healthy compared to
affected tissues.

A common problem in the practical application of graphical models in systems 
biology is the high dimensionality of the data compared to the small sample 
size. In other words, there are a large number $p$ of variables to be considered
whereas the number of observations $n$ is small due to ethical reasons and cost
factors. Typically, the number of parameters in a graphical model grows with
some power of the number of variables. Hence, if the number of genes is large,
the parameters describing the graphical model (e.g. edge probabilities) quickly
outnumber the data points. For this reason graphical modelling in systems biology
almost always requires some form of regularised inference, such as Bayesian 
inference, penalised maximum likelihood or other shrinkage procedures.   

\subsection{Correlation networks}

The simplest graphical models used in systems biology are \textit{relevance
networks} \citep{BTS+00}, which are also known in statistics as \textit{correlation
graphs}. Relevance networks are constructed by first estimating the correlation 
matrix for all $p (p-1)/2$ pairs of genes. Subsequently, the correlation matrix 
is thresholded at some prespecified level, say at $| r_{ij} | < 0.8 $, so that weak
correlations are set to zero. Finally, a graph is drawn in order to depict the 
remaining strong correlations.

Technically, correlation graphs visualise the \textit{marginal} (in)dependence 
structure of the data. Assuming the latter are normally distributed, a missing 
edge between two genes in a relevance network is indicative of marginal stochastic
independence. Because of their simplicity, both in terms of interpretation as 
well as computation, correlation graphs are enormously popular, not only for
analysing gene expression profiles but also many other kinds of omics data 
\citep{Ste06}.

\subsection{Covariance selection networks}

The simplest graphical model that considers conditional rather than marginal
dependencies is the \textit{covariance selection model} \citep{Dem72}, also
known as \textit{concentration graph} or \textit{graphical Gaussian model}
\citep{whittaker}. In a GGM the graph structure is constructed in the same 
way as in a relevance network; the only difference is that the presence of an
edge is determined by the value of the corresponding \textit{partial correlation}
(the correlation between any two genes once the linear effect all other $p-2$
genes has been removed) instead of the marginal correlation used above. Partial
correlations may be computed in a number of ways, but the most direct approach
is by inversion and subsequent standardisation of the correlation matrix (the
inverse of the covariance matrix is often called a \textit{concentration matrix}).
Specifically, it can be shown that if an entry in the inverse correlation matrix
is close to zero then the partial correlation between the two corresponding genes
also vanishes. Thus, under the normal data assumption a missing edge in a GGM
implies conditional independence.

Partial correlation graphs derived from genomic data are often called \textit{gene
association networks}, to distinguish them from correlation-based relevance
networks. Despite their mathematical simplicity, it is not trivial to learn GGMs
from high-dimensional ``small $n$, large $p$'' genomic data \citep{SS05b}. There
are two key problems. First, inferring a large-scale correlation (or covariance) 
matrix from relatively few data is an ill-posed problem that requires some sort
of regularisation. Otherwise the correlation matrix is singular and therefore
cannot be inverted to compute partial correlations. Second, an effective
variable selection procedure is needed to determine which estimated partial 
correlations are not significant and which represent actual linear dependencies.
Typically, GGM model selection involves assumptions concerning the sparsity of
the actual biological network.

The first applications of covariance selection models to genome data were either
restricted to a small number of variables \citep{WK00}, used as a preprocessing step
in cluster analysis to reduce the effective dimension of the model \citep{TH02a},
or employed low order partial correlations as an approximation to fully conditioned
partial correlations \citep{FBHM04}. However, newer inference procedures for GGMs
are directly applicable to high-dimensional data. A Bayesian regression-based 
approach to learn large-scale GGMs is given in \citet{DHJ+04}. \citet{SS05a} 
introduced a large-scale model selection procedure for GGMs using false discovery
rate multiple testing with an empirically estimated null model. \citet{SS05c} also
proposed a James-Stein-type shrinkage correlation estimator that is both 
computationally and statistically efficient even in larger dimensions, specifically for 
use in network inference. An example of a GGM reconstructed with this algorithm
from \textit{E. coli} data is shown in Figure \ref{ecoliggm}. Methods for estimating
large-scale inverse correlation matrices using different variants of penalised 
maximum likelihood are discussed by \citet{LG06}, \citet{BGA2008} and \citet{FHT2008}.
Most recently, \citet{AK2009} considered a modified GGM that allows the specification
of interactions (i.e. multiplicative dependencies) among genes, and \citet{KSB2009}
conducted an extensive comparison of regularised estimation techniques for GGMs.

\begin{figure}[!hp]
  \eps{images/ecoli}{10}
  \caption{Partial correlation graph inferred from \textit{E. coli} data using the
    algorithm described in \citet{SS05a} and \citet{SS05c}. Dotted edges indicate
    negative partial correlation.}
  \label{ecoliggm}
\end{figure}

\subsection{Bayesian networks}

Both gene relevance and gene association networks are undirected graphs. In order
to learn about directed conditional dependencies Bayesian network inference 
procedures have been developed for static (and later also for time course)
microarray data. 

The application of Bayesian networks to learn large-scale directed graphs from
microarray data was pioneered by \citet{FLNP00}, and has also been reviewed more
recently in \citet{Fri04}. The high dimensionality of the model means that inference
procedures are usually unable to identify a single best Bayesian network, settling
instead on a set of equally well behaved models. In addition, as discussed in Section
\ref{sec:structlearn}, all Bayesian networks belonging to the same equivalence class
have the same score and therefore cannot be distinguished on the basis of the 
probability distribution of the data. For this reason it is often important to 
incorporate prior biological knowledge into the inference process of a Bayesian
network. A Bayesian approach based on the use of informative prior distributions
is described in \citet{MS2008}.  

The efficiency of Bayesian networks, GGMs and relevance networks in recovering 
biological regulatory networks have been studied in an extensive and realistic 
setup in \citet{WGH06}. Not surprisingly, the amount of information contained in gene
expression and other high-dimensional data is often too low to allow for accurate
reconstruction of all the details of a biological network. Nonetheless, both GGMs
and Bayesian networks are able to elucidate some of the underlying structure.

\subsection{Dynamic Bayesian networks}
\label{sec:dbns}

The extension of Bayesian networks to the analysis of time course data is provided
by \textit{dynamic Bayesian networks}, which explicitly account for time 
dependencies in their definition. The incorporation of temporal aspects is important
for systems biology, as it allows to draw conclusions about causal relations. 

Dynamic Bayesian networks are often restricted to linear systems, with two special
(yet still very general) models: the vector-autoregressive (VAR) model and state-space
models. The main difference between the two is that the latter includes hidden 
variables that are useful for implicit dimensionality reduction.

The VAR model was first applied to genomic data by \citet{FS+2007} and \citet{OS07b}.
A key problem of this kind of model is that it is very parameter-rich, and therefore it
is hard to estimate efficiently and reliably. \citet{OS07b} proposed a shrinkage approach 
whereas \citet{FS+2007} employed lasso regression for sparse VAR modelling. A refinement
of the latter approach based on the elastic net penalised regression is described 
in \citet{SI+2009}. In all VAR models the estimated coefficients can be interpreted 
in terms of Granger causality \citep{OS07b}.

State-space models are an extension of the VAR model, and include lower-dimensional
latent variables to facilitate inference. The dimension of the latent variables
is usually chosen in the order of the rank of the data matrix. \citet{Hus03}, 
\citet{PR+2003}, and \citet{RA+04} were the first to study genomic data with dynamic
Bayesian networks and to propose inference procedures suitable for use with
microarray data. Bayesian learning procedures are discussed in \citep{LS08}. A 
general state-space framework that allows to model non-stationary time course
data is given in \citet{GH2009}. \citet{RJFD2010} present an empirical Bayes
approach to learning dynamical Bayesian networks and apply it to gene expression data.

\subsection{Other graphical models}

Bayesian networks are graphical models where all edges are directed, whereas GGMs
represent undirected conditional dependencies in multivariate data. On the other
hand, \textit{chain graphs} can include directed as well as undirected dependencies
in same graph.  One heuristic approach to infer an approximating chain graph from 
high-dimensional genomic data is described in \citet{OS07c}.

For reasons of simplicity, and to further reduce the number of parameters to be
estimated, many graphical models used in systems biology only describe linear
dependencies (GGM, VAR, state space models).  Attempts to relax such linearity
assumptions include entropy networks \citep{MKLB07,HS09} and copula-based approaches
\citep{KJ+2008}.

Finally, sometimes time-discrete models such as dynamic Bayesian networks are not
appropriate to study the dynamics of molecular processes. In these cases stochastic
differential equations \citep{Wilkinson2009} often represent a viable alternative.
It is also important to keep in mind that, given the small sample size of omics data,
the most complex graphical model is not necessarily the best choice for an analyst
\citep{WGH06}.

\end{document}